\pdfoutput=1
\documentclass[11pt]{article}

\usepackage[]{ACL2023}

\usepackage{times}
\usepackage{latexsym}
\usepackage{subcaption}
\usepackage[T1]{fontenc}
\usepackage[utf8]{inputenc}

\usepackage{microtype}

\usepackage{inconsolata}
\usepackage{tabularx}
\usepackage{xcolor}
\usepackage{multirow}
\usepackage{multicol}
\usepackage{graphicx}
\usepackage{floatrow}
\usepackage{makecell}
\usepackage{xargs} 
\usepackage{amsthm}
\usepackage{caption}
\usepackage{soul}
\usepackage{color,soul}
\usepackage{mathtools}
\usepackage{array}
\definecolor{pink1}{HTML}{efedf5}
\definecolor{pink2}{HTML}{bcbddc}
\definecolor{pink3}{HTML}{756bb1}

\definecolor{blue1}{HTML}{deebf7}
\definecolor{blue2}{HTML}{9ecae1}
\definecolor{blue3}{HTML}{3182bd}

\usepackage{booktabs}
\usepackage{tabu}
\usepackage{xcolor}
\usepackage{graphicx}

\title{Mitigating Approximate Memorization in Language Models via Dissimilarity Learned Policy}

\author{Aly M. Kassem \\
  University of Windsor\\ Windsor, ON, Canada\\
  \texttt{kassem6@uwindsor.ca} }
\usepackage{background}
\backgroundsetup{
  position=current page.east,
  angle=-90,
  nodeanchor=east,
  vshift=-5mm,
  opacity=1,
  color=gray,
  scale=1.8,
  contents=work in progress
}
\begin{document}
\maketitle
\begin{abstract}
Large Language models (LLMs) are trained on large amounts of data, which can include sensitive information that may compromise personal privacy. LLMs showed to memorize parts of the training data and emit those data verbatim when an adversary prompts appropriately. Previous research has primarily focused on data preprocessing and differential privacy techniques to address memorization or prevent verbatim memorization exclusively, which can give a false sense of privacy. However, these methods rely on explicit and implicit assumptions about the structure of the data to be protected, which often results in an incomplete solution to the problem. To address this, we propose a novel framework that utilizes a reinforcement learning approach (PPO) to fine-tune LLMs to mitigate approximate memorization. Our approach utilizes a negative similarity score, such as BERTScore or SacreBLEU, as a reward signal to learn a dissimilarity policy	$\pi_{D}$. Our results demonstrate that this framework effectively mitigates approximate memorization while maintaining high levels of coherence and fluency in the generated samples. Furthermore, our framework is robust in mitigating approximate memorization across various circumstances, including longer context, which is known to increase memorization in LLMs.
\end{abstract}

\section{Introduction}
Large language models have recently grown exponentially, from millions to billions to trillions of parameters \cite{radford2019language, brown2020language, chowdhery2022palm, fedus2021switch}. As the scale of those models rises, their training sets grow to billions of tokens \cite{gao2020pile}, and their performance improves across the board, even when using a few-shot learning setting \cite{brown2020language}. With the increase in the size of both models and datasets, as well as the performance improvement, practical concerns have arisen regarding the privacy risks of memorization of the training data in large language models, as an adversary interacting with a pre-trained model can extract individual sequences that were used to train the model \cite{carlini2021extracting}, even if the language model was trained on a dataset that is publicly accessible. Many research has been conducted in the context of large language models to address training data memorization problems; studies show that a language model with 6 billion parameters (GPT-J) can memorize at least 1\% of its training data \cite{carlini2022quantifying}. The cause of memorization might be the language models' training strategy, as the objective of a language model \cite{radford2018improving} is to identify the relationship between the present token and its succeeding (auto-aggressive LM) or surrounding segments (MLM). Another cause for memorization might be several repeated instances in the training corpus since the more repeated an example is, the more likely it is to be memorized \cite{lee2021deduplicating}. 

% Several approaches have been proposed to address the problem of memorization in large language models. One approach is data sanitization, which involves removing private information from the training data \cite{dernoncourt2017identification, lison2021anonymisation}. Another approach is the use of differential privacy (DP) algorithms \cite{wang2019beyond, mcmahan2017learning}, which are designed to prevent the memorization of private data. Data deduplication \cite{kandpal2022deduplicating} is another approach that can reduce the risk of memorization by removing duplicated content in the training corpus. While effective, this approach does not guarantee that the model will not memorize other instances, as not all memorized data are duplicated.
% Additionally, applying a defense at inference time \cite{ippolito2022preventing} can effectively prevent the generation of memorized content, not only for verbatim memorization (exact matching of a substring from the training set) but also for approximate memorization. However, one disadvantage of these techniques is that the data sanitization approach assumes that private information can be formally characterized, discrete, and context-independent, which is only sometimes the case as private information is often context-dependent, not easily identifiable, and not discrete. Differentially private training also has the drawback of resulting in lower-quality generative models\cite{anil2021large}.

Several approaches have been proposed to address the problem of memorization in large language models, such as data sanitization, the use of differential privacy algorithms, and data deduplication. These techniques can effectively prevent the generation of memorized content, but they also have drawbacks. For example, data sanitization assumes that private information can be easily identified and is not context-dependent, while differential privacy can result in lower-quality generative models \cite{anil2021large}.

In this study, we propose a framework to prevent memorization in large language models by fine-tuning those models using a reinforcement learning approach (PPO) \cite{schulman2017proximal}. Given samples of prefixes and suffixes from the original pre-training data of the language model, we use a prefix as input for the language model to generate the suffix; then, we compute the negative SacreBLEU \cite{post-2018-call} score to measure the dissimilarity between the true suffix and generated suffix, the dissimilarity scores is then regarded as a reward signal to maximize in the training process, which guarantees that the 
approximate memorization will be mitigated. We experimented with different reward functions such as BERTScore \cite{zhang2019bertscore}, the weighted sum between perplexity, and SacreBLEU or BERTScore. The objective of those experiments is to see how they can affect the generation quality and memorization ratio. Yet, all of the proposed reward functions assure to minimization of approximate memorization. The aim of the framework is to learn a policy $\pi_{D}$ that can paraphrase the suffix given some prefix. For example, in” Alice green lives at 187 bob street”, the prefix is ”Alice green lives at.” The suffix is ”187 bob street” we aim that the fine-tuned language model paraphrase the suffix to be: 12 red street. As the suffix is paraphrased, the memorization relationship between the prefix and suffix is minimized. 
We evaluate the effectiveness of the framework in two different settings. We consider the first one as the training process's standard setting, giving a prefix to the model; we want the generated suffix to be as dissimilar as possible to the true suffix. Many studies \cite{carlini2021extracting, carlini2022quantifying} have shown that as a longer context is provided, the memorization ratio increases, so the second setting is to provide 100 additional context tokens pre-prefix combined with the prefix to the model to evaluate the memorization in this case. The proposed framework does not make any explicit or implicit assumptions about the structure of the data to be protected. Also, unlike the DP methods, the proposed framework does not apply any partition mechanism to split the data into public data and private data; as language data cannot be partitioned, we apply the policy on all training data as defining or partitioning the data into private and public might be impossible in case of language data\cite{brown2022does}.

We tested with three model sizes (125M, 1.3B, and 2.7B parameters), and all of our models are the GPT-Neo Family\cite{gpt-neo}. Our main findings are as follows:
\begin{itemize}
    \item \textbf{The learned policy is able to generate new suffixes which are different from the true ones} by a large margin of dissimilarity without a considerable loss in the generation quality that's  achieved because the fine-tuned LM learned a policy to change names, numbers, or replace a whole phrase by a similar entity.

    \item \textbf{As the size of the language model increases, the convergence rate becomes faster}. The convergence in this context means that the model-generated suffixes become significantly different from the original suffixes, and the difference between the perplexity of the generated examples and the original examples becomes smaller. In our experiments, we found that the GPT-Neo 125M converged in three epochs, four PPO epochs per
    batch, GPT-Neo 1.3B model converged in two epochs, four PPO epochs per batch, and the GPT-Neo 2.7B model converged in two epochs, two PPO epochs per batch.
    \item \textbf{As the size of a language model increases, the dissimilarity score increase}, which can be measured by the difference between negative SacreBLEU before and after applying the framework. However, this increase in dissimilarity score is only sometimes a positive outcome. In some settings, it is accompanied by an increase in the perplexity score of the generated examples. This suggests that larger models may tend to "forget" the memorized data faster.
\end{itemize}
Overall, our findings show that using the proposed framework to fine-tune large language models mitigates training data approximate memorization while ensuring suffix generation quality without a considerable loss in the fluency and coherence of the text.

\section{Background}
\subsection{Language Models}
Language models are central to natural language processing techniques. They operate by taking in a sequence of tokens, such as words or characters, and outputting a probability distribution over the next token. Training a language model aims to maximize the likelihood of a given text corpus. One popular approach to training language models is to use a "next-token prediction" objective, which involves constructing a generative model of the probability distribution of a sequence of tokens. State-of-the-art language models use neural networks, such as recurrent neural networks or attention-based models, to estimate this probability distribution. In recent years, transformer-based language models \cite{vaswani2017attention, devlin2018bert, radford2019language} have become particularly popular due to their ability to scale to billions of parameters and achieve high performance on large datasets, as it has been proved by empirical results that show that the performance of transformer-based language models follows a power law relationship concerning model size, dataset size, and amount of compute used for training \cite{kaplan2020scaling}. This means that to see significant improvements in model performance, there needs to be an order-of-magnitude increase in at least one of these factors. For example, increasing the model size by a factor of 10 or increasing the dataset size by a factor of 10 could lead to a noticeable improvement in model performance. It is worth noting that using large models, large datasets, and high amounts of compute time are all essential for achieving high performance with language models. However, these large language models have also raised concerns about their potential to memorize and replicate sensitive information, such as personal information or long text sequences \cite{carlini2021extracting, brown2022does}. In our experiments, we considered three different auto-regressive large language models: GPT-Neo 125M, GPT-Neo 1.3B, and GPT-Neo 2.7B, which are pre-trained on English corpora. 
\subsection{Memorization Definitions}
In the context of memorization in large language models, we follow the definition proposed by \cite{lee2021deduplicating}, which introduced approximate memorization. Given a string S, if a prompt P exists, the model generates ‘s’ given ‘P,’ and the model output is memorized with some chosen edit distance of the prompt’s true continuation in the training set. In our study, we choose the edit distance to be a similarity measure (BLEU) as proposed in \cite{ippolito2022preventing}, to be able to capture the approximate memorization, not just the “Eidetic memorization” \cite{carlini2021extracting} as the definition of verbatim memorization fails to include more subtle forms of memorization \cite{ippolito2022preventing}. For example: if we have a sequence S: “My name is Alice green, I live at 187 bob street, and my phone number is 226284”, and the model generation is: “My name is Alice green, I live at 187 queen street, and my phone number is 226284” following the definition of “verbatim memorization,” this sequence isn’t memorized, but if we follow “approximate memorization” we can say that the model generation is similar by 84.92\% SacreBLEU score to the true continuation. 

\subsection{Related Work}
Recent studies \cite{ippolito2022preventing} reveal that large language models may avoid filters that limit verbatim memorization by performing plausible "style transfers" to the prompt. They demonstrated this by proposing an inference time defense mechanism, "MemFREE decoding," to prevent memorization in large language models. This is accomplished by efficiently querying the training dataset and checking for the existence of any n-gram in the prefix combined with the generated suffix. Another study showed that duplicate instances in the training data increase the likelihood that a language model will memorize them. So data deduplication techniques \cite{kandpal2022deduplicating} effectively remove these duplicates from the training data. However, it is essential to note that this approach only partially prevents the model from memorizing instances, as it is still possible to memorize sequences beyond duplicated instances.
Moreover, Differential privacy (DP) is a widely-used technique for training models that do not memorize individual training examples. DP algorithms, such as the ones described in \cite{abadi2016deep}, are considered the gold standard for protecting privacy in machine learning models. However, in practice, these techniques often result in worse performance than non-private models, as observed by \cite{anil2021large}. As a result, state-of-the-art language models often require a large amount of data and computational resources to train and are not typically trained using DP. Furthermore, DP algorithms are computationally expensive, with slower convergence and lower utility than non-private methods \cite{anil2021large}. This is especially concerning for language models, which are known to memorize large portions of their training data and may be more likely to memorize anomalous data points, presenting a privacy risk for the text's authors or subjects. One of the most challenging aspects of using DP on language data is identifying the boundaries of private information \cite{brown2022does}.

\section{Data}
We employed a subset of the Pile dataset, which was released as a benchmark for training data extraction attacks on large Language Models. Generally, the Pile dataset contains data from various sources (e.g., books, Web scrapes, open source code). We used this version of the subset \footnote{\url{https://github.com/google-research/lm-extraction-benchmark}}, designed to be easy to extract to assess the performance of targeted attacks. The dataset contains only 15,000 samples since the full version is not released yet. Each sample consists of 200 tokens sampled randomly from the Pile training set. The topics included in the subset are code, news, logs, conversations-replies, copyrights, links, etc. Most of them are in the English language. The dataset is splitted into 13,500 samples for training and 1,500 samples for testing.\newline

\textbf{Training Data.}
For the training phase, we use the third 50 tokens for each string as a prefix and the fourth 50 tokens as a suffix.\newline

\textbf{Evaluation Data.}
We employed various settings and datasets to ensure the learned policy generalization. First, we evaluated the fine-tuned model on the test set of the Pile-subset as we use the third 50 tokens for each string as a prefix and the fourth 50 tokens as a suffix. Second, in the longer context setting, we added extra 100 tokens and combine it with the third 50 tokens in the sample; the first 100 tokens represent the pre-prefix, and the next 50 tokens are prefixes. Using a longer context in a language model can be considered a form of attack, as it allows the adversary to gain access to more information. With this additional information, the adversary can extract new and sensitive information from the language model, compromising its security and integrity.

\begin{figure}[]
\centering
\includegraphics[width=0.45\textwidth]{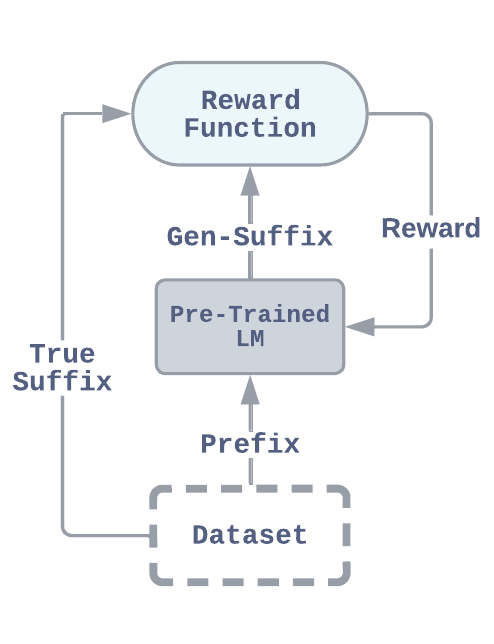}

\caption{Illustration of Framework Pipeline which mitigates approximate memorization in language models}
\label{fig:lm_pipeline}
\end{figure}

\section{Method}
Our approach begins with an autoregressive pretrained language model (GPT-Neo 125M, GPT-Neo 1.3B, and 2.7B), a dataset of 15,000 samples split into prefixes and suffixes, and a reward function (e.g., Negative SacreBLEU), which will be discussed later. The following steps are followed: Feed the prefixes into the pretrained language model to obtain the generated suffixes. We pass the generated suffixes to the reward function, which computes the dissimilarity between the generated and true suffixes. Finally, we use the reward function output as a scaler reward to fine-tune the language model and optimize the reward using the PPO algorithm.

\subsection{Training Environment}
We employed GENERATION AS A TOKEN-LEVEL MDP environment \cite{ramamurthy2022reinforcement}; the environment, in general, is an NLP task in which we have a supervised dataset consisting of prefixes and suffixes. Generating a response can be thought of as following a Markov Decision Process, where each step involves choosing a vocabulary word as an action based on the current input and previous actions. Each episode starts with a specific prompt and ends when a certain number of steps have been taken(in our case, the number of steps means some tokens that have been generated) or an end-of-sentence token is generated. The reward for an episode is based on how well the final state is dissimilar to the target output (e.g., an automated metric like Negative SacreBLEU, BERTScore).

\subsection{Fine-tuning details}
Using PPO, we fine-tuned the pretrained language model in our environment. The environment used is GENERATION AS A TOKEN-LEVEL MDP, similar to the bandit environment in that it presents a random customer prompt and expects a response. The only difference between the employed and bandit environments is that in the bandit, we utilize a discount factor $\gamma = 0.95$ instead of $\gamma = 1$. The reason for selecting $\gamma= 0.95$ is that it gives more stability in training since the rewards are calculated in a discounted fashion in the token-level MDP, which reduces the magnitude of the reward that is applied to tokens selected at the start \cite{ramamurthy2022reinforcement}. Given the prefix and generated suffix, it computes a reward based on the reward function of choice, and the episode is completed.
Furthermore, the KL penalty is applied per token using a reference model (the pretrained model before fine-tuning.) This prevents the fine-tuned model from generating a suffix that deviates too much from the reference language model (e.g., generating white spaces). A value network $V$ is included beside the language modeling head to estimate the value function. The batch size is 32 for all models; we selected a specific number of epochs for each model as the convergence rate for each model is different; we mean by convergence in this context that the model-generated suffixes become significantly different from the original suffixes but without a considerable loss in the perplexity as the difference between the perplexity of the generated examples and original examples becomes smaller, so we selected the appropriate number of epochs that balance between these goals. We use three epochs, four PPO epochs per batch for GPT-Neo 125M, two epochs, four PPO epochs per batch for GPT-Neo 1.3B, and two epochs, two PPO epochs per batch for GPT-Neo 2.7B. The learning rate for all models was $1.41 \times 10^{-5}$. Also, the KL Beta of 0.2 is selected, and the clip range of 0.2.
\subsection{Reward Functions}
To effectively prevent the approximate memorization of training data and encourage learning a diverse set of language generation strategies, we must carefully design a suitable reward function for our pretrained language model. This reward function should enable the model to generate dissimilar suffixes that are semantically consistent, have the correct syntactic structure, and are related to the same topic as the prefix. Previous research has shown that the pretrained LM can bypass memorization checks by producing ALL-CAPITAL text \cite{ippolito2022preventing}, so our reward function must address this issue. We conducted experiments with three different reward functions based on the semantic similarity between the model's generated text and the training data. However, we observed that the model could learn a policy to maximize rewards in ways that do not align with the desired output, such as generating white space or repeating the same words multiple times through learning shortcuts \cite{geirhos2020shortcut}. Our chosen reward function must address these potential pitfalls.\newline

 \begin{table*}[htbp]

\begin{center}
    \centering
    \captionsetup{justification=centering}
\caption{The Results Of Different Reward Functions On The Standard Setting Test Set.}

\label{table:rf}
\small
\resizebox{9.2cm}{!}{%

\centering
\begin{tabular}{@{}c  c  c  c  c  c @{}}
\toprule

\textbf{Model} & \textbf{Reward Function}  & \textbf{Epoch} & $\bf N-SacreBLEU\uparrow$ & $\bf PPL\downarrow$\\\midrule

\multirow{3}{*}{GPT-Neo 125M} & N-SacreBLEU
 & \multirow{3}{*}{12} & 82.78 & 5.25\\
 & N-SacreBLEU + PPL & &78.32&4.83\\
& \textbf{BERTScore} & & $\bf 67.70$ & $\bf 4.01$
\\\midrule

\multirow{3}{*}{GPT-Neo 1.3B} & N-SacreBLEU
 & \multirow{3}{*}{8} & 72.10 & 3.652\\
 & N-SacreBLEU + PPL & &52.02&2.85\\
& \textbf{BERTScore} & & $\bf 53.22$ & $\bf 2.64$

\\\midrule

\multirow{3}{*}{GPT-Neo 2.7B} & N-SacreBLEU
 & \multirow{3}{*}{4} &71.47 & 6.105 & \\
 & N-SacreBLEU + PPL & &80.84&8.60\\
& \textbf{BERTScore} & & $\bf 61.79$ & $\bf 4.46$
\\\midrule

\end{tabular}%
}
\normalsize
\end{center}
\end{table*}

\textbf{Negative SacreBLEU.} is a popular metric for evaluating the quality of the machine-translated text, which is a modified version of the BLEU (Bilingual Evaluation Understudy) metric. Our approach uses negative SacreBLEU, calculated as the difference between 100 and the SacreBLEU score. Negative SacreBLEU has the advantage of being based on semantic similarity, making it suitable for learning policies that minimize memorization. However, it has the limitation of being restricted to a maximum n-gram length and not considering contextual meaning, which may limit the model’s ability to learn diverse and robust policies. In the following sections, we will explore ways to address this limitation.

\textbf{Negative SacreBLEU + Perplexity.} To improve the fluency of the model's generated text, we experimented with combining the Negative SacreBLEU score with the perplexity metric. We found that as the dissimilarity score (measured using Negative SacreBLEU) increased, the fluency of the model (as measured by perplexity) decreased. Therefore, we incorporated perplexity as a penalty term in the overall reward function, which was calculated as the weighted sum of Negative SacreBLEU and the negative of the perplexity metric, with a weight of 0.5 applied to each. This combination of metrics allowed us to balance the goal of encouraging dissimilarity with the need to maintain fluency in the model's generated text.

\textbf{BERTScore.} is a model-based semantic metric for assessing the quality of text generated by a model. It works by embedding the text using a BERT-based model and calculating the cosine similarity between the two resulting embedding vectors. BERTScore outputs recall, precision, and F1 scores based on this similarity. In our experiments, we employ BERTScore computed by DeBERTa large model \cite{he2021deberta} to maximize the negative F1-score, which can be interpreted as minimizing the cosine similarity between the generated and true suffix. One advantage of using BERTScore is that it relies on contextualized embeddings, which allows it to capture dependencies and consider the contextual meaning of words in the sentence. This can enable the model to learn more diverse and contextually appropriate policies.

\section{Experiments}
To thoroughly assess the effectiveness of our framework for preventing memorization in large language models, we conducted a series of experiments using different settings and evaluation metrics. We evaluated the model in standard settings and longer contexts. Our evaluations covered a range of factors, including dissimilarity measures, semantic consistency, and fluency. This comprehensive approach allowed us to analyze the model's performance in various scenarios and understand its capabilities and limitations.

\subsection{Experimental Setting}
To evaluate the performance and flexibility of our framework, we conducted a series of experiments using different settings and test sets. These experiments included variations in the model size, reward function, and the number of epochs. This allowed us to analyze the impact of these factors on the model's performance and identify the optimal configuration for the task.

\textbf{Models.} To investigate the impact of model size on our framework and its feasibility for different sizes, we conducted experiments using three models with varying sizes: 125M, 1.3B, and 2.7B. These models are part of the GPT-Neo family, and the memorization problem was observed using these models \cite{carlini2022quantifying}. The fact that they have been subjected to various techniques for extracting or preventing memorization in the literature makes them a benchmark to study the memorization problem. An additional advantage of the GPT-Neo models is that their training set is known; they were pretrained on the publicly available Pile dataset, which consists of 825GB of data.

\textbf{Reward Functions.} To examine the effect of different reward functions on the learned policy in our proposed framework, we conducted experiments using three different reward functions: negative SacreBLEU, negative SacreBLEU combined with perplexity, and negative BERTScore. These reward functions were designed to address the issue of approximate memorization and encourage the generation of dissimilar yet semantically consistent output. They are based on the notion of semantic similarity, which ensures that the generated text is meaningfully related to the input. By comparing the performance of these reward functions, we aimed to identify the optimal choice for the given task.

\textbf{Number of Epochs.} The number of epochs is a key hyperparameter in our framework, as it determines the number of times the model is trained on the data. During each epoch, the learned policy can change, so it is important to carefully select the number of epochs in order to find the optimal policy. There is a tradeoff between the dissimilarity and fluency of the generated examples, which we will discuss later. Therefore, choosing the appropriate number of epochs is crucial for achieving the best balance between these competing objectives.

\textbf{Dataset Settings.} To evaluate the ability of the learned policy to generalize to different settings in the pretraining dataset, we conducted experiments using three different context lengths: the same training sequence length and an additional 100 tokens as a longer context. This allowed us to assess how the model's performance is affected by the length of the context and to identify any patterns or trends in its behavior. By comparing the results across these different context lengths, we aimed to understand the model's generalization capabilities better.

\textbf{Evaluation Metrics.} To assess the performance of our method, we evaluated two key aspects: the dissimilarity score and the quality of the generated suffixes.
(1) Dissimilarity Score: We measured the dissimilarity between the generated suffix and the true suffix using the negative SacreBLEU score, as it is based on semantic similarity and can, therefore, effectively measure approximate memorization.
(2) Quality of generated suffixes: To assess the text's fluency, we used the perplexity score of the applied model before applying the framework. This metric allowed us to evaluate the quality of the generated suffixes in terms of their grammatical correctness and overall coherence. We could comprehensively understand the model's performance by considering both the dissimilarity score and the quality of the generated suffixes.

\begin{figure}
     \centering
     \begin{subfigure}[b]{0.35\textwidth}
         \centering
        \fbox{{\includegraphics[width=2.8cm, height=2.4cm]{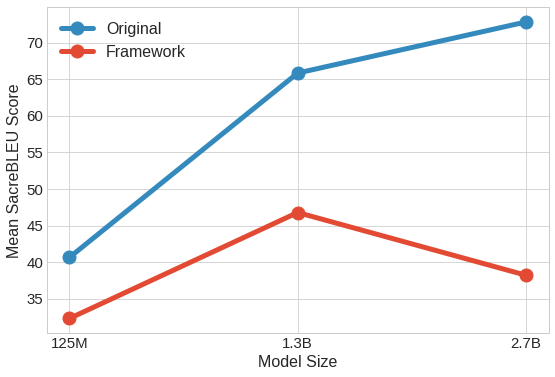}}} 
         \caption{Standard Setting}
         \label{fig:y equals x}
     \end{subfigure}
     \begin{subfigure}[b]{0.6\textwidth}
         \centering
         \fbox{{\includegraphics[width=2.8cm, height=2.4cm]{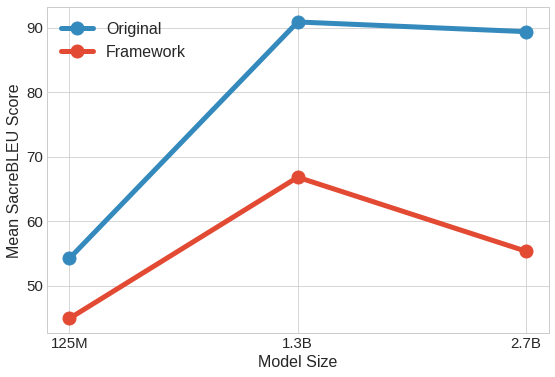}}} 
         \caption{Longer Context Setting}
         \label{fig:three sin x}
     \end{subfigure}
        \caption{Comparing The Mean SacreBLEU Score Before \textcolor{blue}{(blue)} \& After Applying the Framework \textcolor{orange}{(orange)} Across All Model Sizes}
        \label{fig:lineplot}
\end{figure}

\begin{table*}[]
\centering
\resizebox{9.4cm}{!}{%
\begin{tabular}{lccc|ccccc}
\toprule
\multirow{2}{*}{\textbf{Model}} & 
\multirow{2}{*}{\textbf{Setting}}&
\multicolumn{2}{c}{\textbf{BEFORE}} & \multicolumn{2}{c}{\textbf{AFTER}} &
\multicolumn{1}{c}{\textbf{$\bf \Delta_{N-SB}$}}
&\multicolumn{1}{c}{\textbf{Epoch}}
\\ \cline{3-6} 
    &  &$\bf N-SacreBLEU\uparrow$  & $\bf PPL\downarrow$ &  $\bf N-SacreBLEU\uparrow$  & $\bf PPL\downarrow$   \\ \toprule 
    
\multirow{2}{*}{GPT-Neo 125M} & 
Standard & $59.63$ & $3.85$ & $67.70$& $4.01$  
 & \colorbox{pink1}{8.07} &\multirow{2}{*}{$12$} \\
& LC\textsuperscript{$\ddagger$} &45.74 & 4.12 &  55.04 & 4.15 & \colorbox{blue1}{9.3}

\\ \toprule
 
\multirow{2}{*}{GPT-Neo 1.3B} & 
Standard & $34.16$ & $2.18$ & $53.22$& $2.64$  & \colorbox{pink2}{19.09}&\multirow{2}{*}{$8$}  \\
& LC &9.11 &  1.55 &  36.16 & 1.71 & \colorbox{blue2}{27.05}

% $53.22$& $2.64$ &$8$

\\ \toprule

\multirow{2}{*}{GPT-Neo 2.7B} & 
Standard & $27.18$ & $1.92$ & $61.79$& $4.46$  & \colorbox{pink3}{34.61}&\multirow{2}{*}{$4$} \\
& LC &10.61 &  1.41 &  44.62 & 1.82 & \colorbox{blue3}{34.01}

\\
\toprule
% \multicolumn{17}{l}{}.
\end{tabular}%
}
\caption{Comparsion Of Negative SacreBLEU \& Perplexity Means Before \& After Applying The Framework. \textsuperscript{$\ddagger$} Refer to Longer Context Setting. $\bf \Delta_{N-SB}$ For The Difference Between Negative SacreBLEU Means Before \& After Applying The Framework.}
\label{table:comps}
\end{table*}

\subsection{Experimental Results}
This section will present the results of using the proposed framework with various configurations on the selected dataset. The objective is to examine the effectiveness of the proposed approach under different scenarios and to identify any patterns or trends that may emerge.

\textbf{GPT-Neo 125M.} In our experiments, we found that the negative BERTScore reward function outperformed negative SacreBLEU and negative SacreBLEU combined with Perplexity in terms of achieving a balance between similarity and fluency when used with the GPT-Neo 125M model and a range of epochs from 4, 8, and 12. While negative SacreBLEU combined with the Perplexity achieved a higher dissimilarity score, it came at the cost of a decrease in the fluency of the generated examples as the perplexity score increased. The worst performance was seen with the negative SacreBLEU reward function, as it had a higher dissimilarity score and perplexity score than the other reward functions, as shown in \autoref{table:rf}. Upon further exploration, we found that 12 PPO epochs achieved the best balance between similarity and fluency for all reward functions. All of these experiments were evaluated on the standard test set, and the top three experiment settings were then evaluated on the longer context setting. Negative BERTScore with 12 PPO epochs achieved the best results in this setting. As expected, the longer context increased the memorization score. However, the dissimilarity score also significantly improved, going from 45.74\% before the framework was applied to 55.04\% with the negative SacreBLEU score. As demonstrated in \autoref{table:comps}, the standard setting and the longer context setting results show that the framework is robust in improving the language model's performance. In the standard setting, the difference between negative SacreBLEU before and after applying the framework is 8.07, indicating a significant improvement. Even in the longer context setting, where the model has access to more information, the difference between negative SacreBLEU before and after applying the framework is 9.3, which is still an improvement. This demonstrates that the framework is able to maintain its effectiveness even when the model is presented with more information, which can be a challenge for many models as it increases the risk of memorization. Moreover, using the longer context positively impacted the generated suffixes as the fluency was enhanced.

\begin{figure}[]
\centering
{\frame{\includegraphics[width=0.74\textwidth]{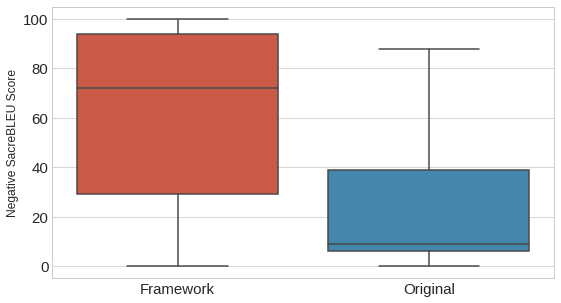}}}

\caption{Displaying The Negative SacreBLEU Distribution Of GPT-Neo 2.7B On Standard Setting Before \textcolor{blue}{(blue)} \& After \textcolor{orange}{(orange)} Applying The Framework}
\label{fig:boxplot}
\end{figure}

\textbf{GPT-Neo 1.3B \& GPT-Neo 2.7B.} We conducted the same experiments with various settings on the GPT-Neo 1.3B and GPT-Neo 2.7B models and found that the same settings performed the best, except for the number of epochs. The GPT-Neo 1.3B model converged at two epochs, four PPO epochs per batch, while the GPT-Neo 2.7B model converged at two epochs, two PPO epochs per batch. This suggests that larger models converge faster. We noticed the same observation of using a longer context which improved the generated suffixes by enhancing fluency and increased the dissimilarity score from 9.11\% to 36.16\% and 10.61\% to 44.62\% of the negative SacreBLEU score in GPT-Neo 1.3B and GPT-Neo 2.7B respectively. However, in those models, the difference increased, which shows that increasing the model size led to an increase in the difference of dissimilarity score as shown in \autoref{fig:lineplot} and, in the case of a longer context setting without a significant loss in perplexity, as the perplexity score increased from 1.55 to 1.71 and 1.414 to 1.823 in GPT-Neo 1.3B and GPT-Neo 2.7B respectively after applying the framework as shown in \autoref{table:comps}.

\section{Discussion}
\textbf{Learned Policy.} We monitored the model's performance as we trained it over different epoch ranges. In an attempt to increase the dissimilarity between the generated and true suffixes, the model initially employed a policy of outputting the same suffix but with different cases (e.g., all uppercase or all lowercase). However, this approach was not useful for our objective and did not reduce similarity. Over epochs, the model improved its learned policy to achieve a better reward. Initially, it tried to minimize the similarity between the target and generated suffixes by altering individual words or numbers. However, this approach could have been more effective in reducing similarity. As training continued, the model began to modify multiple words in the generated suffix in an attempt to increase dissimilarity. Eventually, it started rephrasing or replacing entire phrases with semantically similar alternatives in order to achieve the desired dissimilarity score. It's worth noting that while the dissimilarity score improved with the number of epochs, the perplexity score also increased, indicating a trade-off between the two objectives. Additionally, training the model for too many epochs caused it to generate suffixes that were semantically dissimilar to the true suffix and, in some cases, generated meaningless tokens like question marks or repeated the same word multiple times.\newline
\textbf{Evaluating Approximate Memorization.} According to recent research, when requiring a binary label, whether approximate memorization occurred or not, the BLEU score of 75\% for the generated suffix is the suitable threshold. This threshold was determined by examining a large number of generated examples \cite{ippolito2022preventing}. However, in our own investigation, we found that this issue is mitigated even when the threshold is as low as 50\% after applying the framework. Despite this discovery, we chose to use the more widely accepted threshold of 75\% in order to demonstrate the effectiveness of our framework. After implementing the framework with GPT-Neo 1.3B and GPT-Neo 2.7B, the number of generated examples that exhibited approximate memorization decreased from 910 to 497 and from 1036 to 321, respectively, as shown in \autoref{fig:sp_125m}, \autoref{fig:sp_1.3b}, and \autoref{fig:sp_2.7b}. \autoref{fig:boxplot} shows the negative SacreBLEU scores of GPT-Neo 2.7B on standard setting for both the generated and true suffixes in the form of a box plot. The median negative SacreBLEU score for the generated suffixes is 72.06\%, which means that approximately 50\% of the data has a dissimilarity score higher than 72.06\%. On the other hand, the median negative SacreBLEU score for the true suffixes is 8.70\%, indicating that approximately 50\% of the data has a dissimilarity score lower than 8.70\%. This comparison demonstrates the effectiveness of our framework in reducing the occurrence of approximate memorization.
\section{Conclusion}
In this paper, we propose a novel framework for addressing the issue of large language models memorizing training data. Our approach is demonstrated to be effective through a series of evaluations conducted in various settings. We show that our framework is able to effectively reduce approximate memorization by significantly decreasing the SacreBLEU score while maintaining the fluency and coherence of generated samples. Additionally, our framework shows robustness when using a longer context, which can be seen as a form of attack. While many studies have established a correlation between increasing the size of language models and their tendency to memorize more training data, also, finetuning these larger models can be computationally costly. Our proposed framework offers a new approach to this problem by demonstrating that as the size of the language model increases, the convergence rate and dissimilarity score also increase. These improvements effectively mitigate the issue of increased memorization while reducing the computational costs associated with finetuning larger models. 

\section*{Limitations}
One of the limitations of our work is that it relies on a single scalar reward for optimization, as the problem has dual objectives: dissimilarity and perplexity. To overcome this limitation, we suggest exploring other techniques, such as Multi-objective Reinforcement Learning, which can potentially enhance performance and optimize both objectives simultaneously. Additionally, the dataset used in our work lacks metadata and labels, which can be useful for further analysis of the model's performance on different types of text, such as personal information, copyrights, and news. Using such metadata and labels can help to understand the model's performance on different classes of text and make necessary adjustments to improve the model's performance.
\section*{Ethics Statement}
Improving the large language model to be privacy-preserving is crucial since the language models have become more prominent and involved in many applications in multi-aspect of life. Ensuring the data privacy of those models is vital since some adversary may be able to reach that information. To make those models widely used, we have to guarantee they cannot emit private data. In this paper, we hope our work will serve as a foundation for developing new and innovative solutions to the problem of approximate memorization in large language models since verbatim memorization can give a false sense of privacy, as earlier work suggested. Our proposed framework provides a promising approach to addressing this issue. Further research and experimentation in this area can lead to even more effective methods for reducing memorization in these models. Our work also highlights the importance of considering both the computational cost and the performance trade-off when developing new techniques for addressing memorization in large language models.

% \section*{Acknowledgements}

\bibliography{anthology,custom}
\bibliographystyle{acl_natbib}
\clearpage

\onecolumn
\appendix

\section{Displaying Approximate Memorization Threshold}
\label{sec:appendix}
Recent studies suggested that approximate memorization occurs at the BLEU score of 75\%; we follow this suggestion and demonstrate the effectiveness of the proposed framework in this section by comparing the number of samples that exceed this threshold before and after applying the framework.
\begin{equation}
    \text{SacreBLEU(suffix$_{G}$, suffix$_{T}$)} > 0.75
\end{equation}

\begin{figure}[ht] 
  \begin{subfigure}[b]{0.5\linewidth}
    \centering
    \fbox{\includegraphics[width=0.9\linewidth]{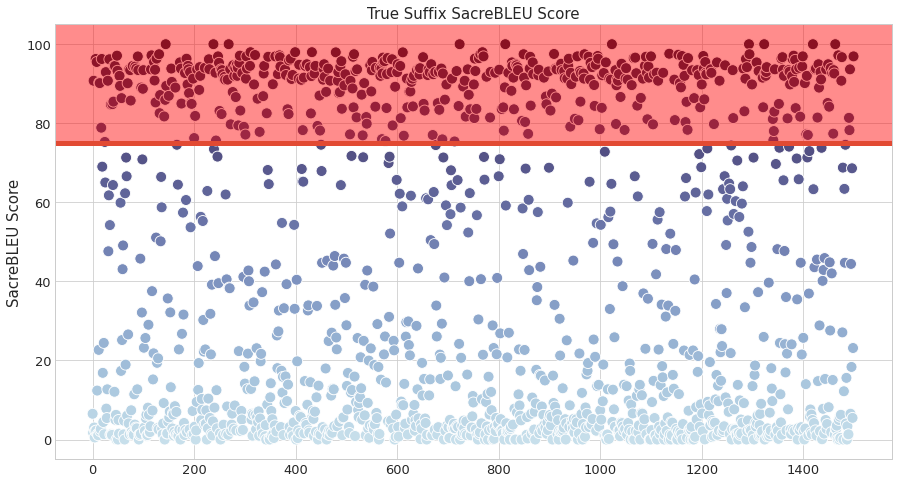}}
    \caption{True Suffixes Standard Setting} 
    \label{fig7:a} 
    \vspace{4ex}
  \end{subfigure}%% 
  \begin{subfigure}[b]{0.5\linewidth}
    \centering
    \fbox{\includegraphics[width=0.9\linewidth]{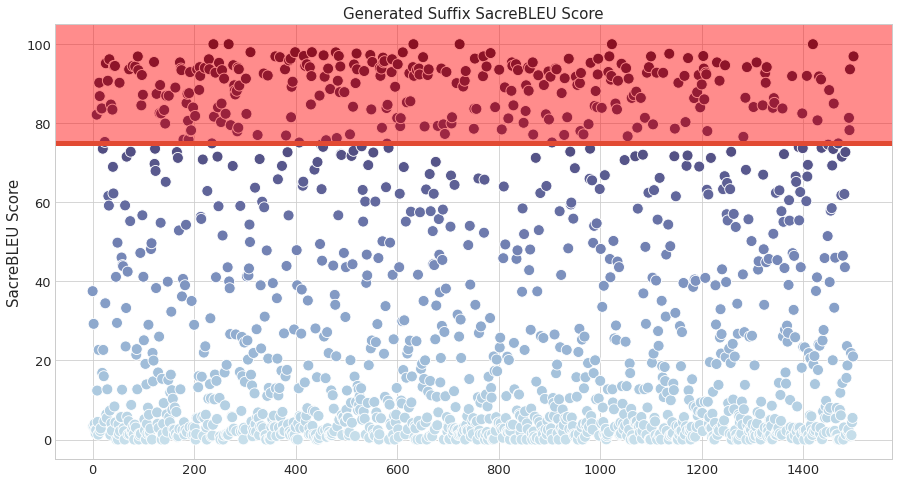}}
    \caption{Generated suffixes Standard Setting} 
    \label{fig7:b} 
    \vspace{4ex}
  \end{subfigure} 
  \begin{subfigure}[b]{0.5\linewidth}
    \centering
    \fbox{\includegraphics[width=0.9\linewidth]{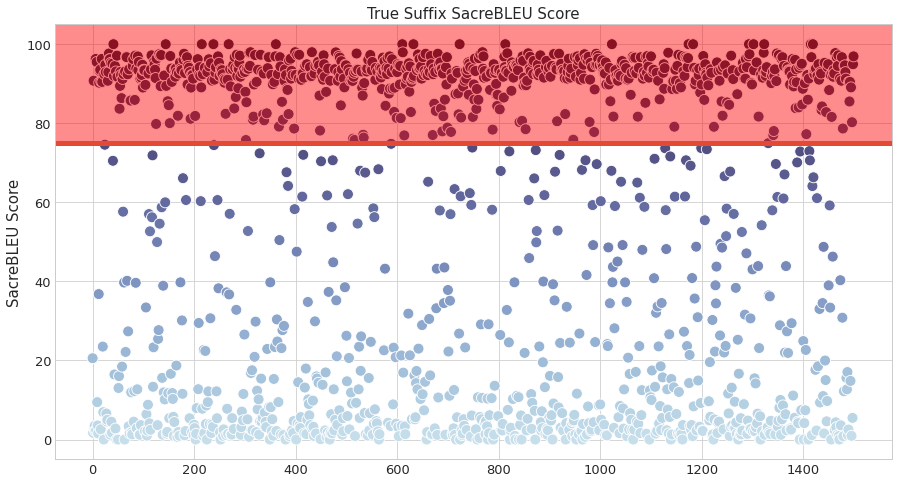}} 
    \caption{True Suffixes Longer Context Setting} 
    \label{fig7:c} 
  \end{subfigure}%%
  \begin{subfigure}[b]{0.5\linewidth}
    \centering
    \fbox{\includegraphics[width=0.9\linewidth]{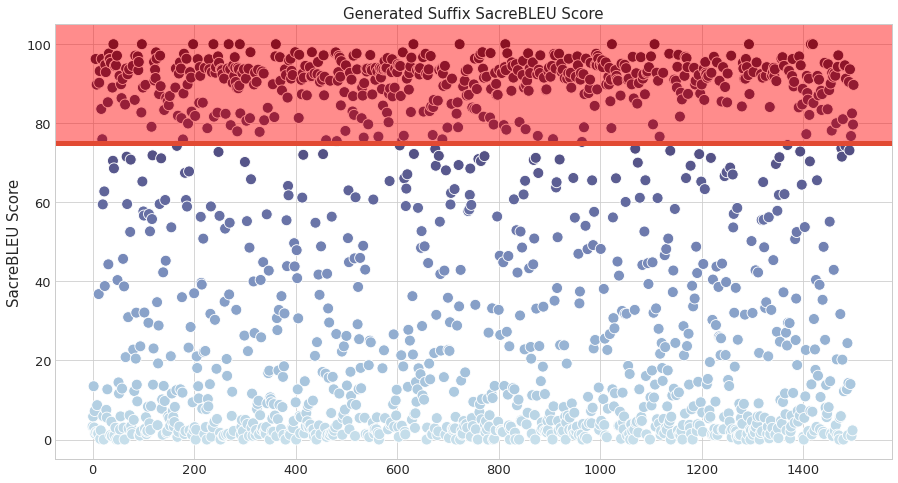}}
    \caption{Generated Suffixes Longer Context Setting} 
    \label{fig:sads} 
  \end{subfigure} 
  \caption{Threshold of 75\% Of The True \& Generated Samples SacreBLEU For GPT-Neo 125M Standard Setting}
  \label{fig:sp_125m} 
\end{figure}
\begin{figure}[ht] 
  \begin{subfigure}[b]{0.5\linewidth}
    \centering
    \fbox{\includegraphics[width=0.9\linewidth]{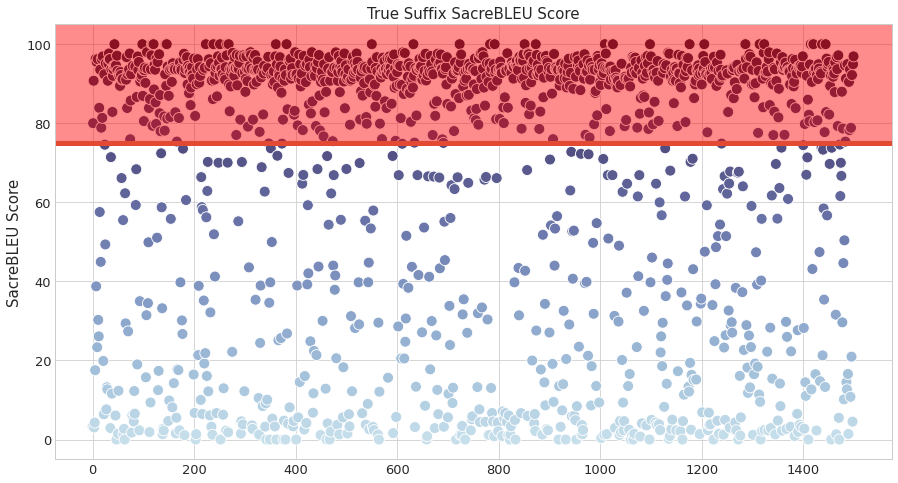}}
    \caption{True Suffixes Standard Setting} 
    \label{fig7:a} 
    \vspace{4ex}
  \end{subfigure}%% 
  \begin{subfigure}[b]{0.5\linewidth}
    \centering
    \fbox{\includegraphics[width=0.9\linewidth]{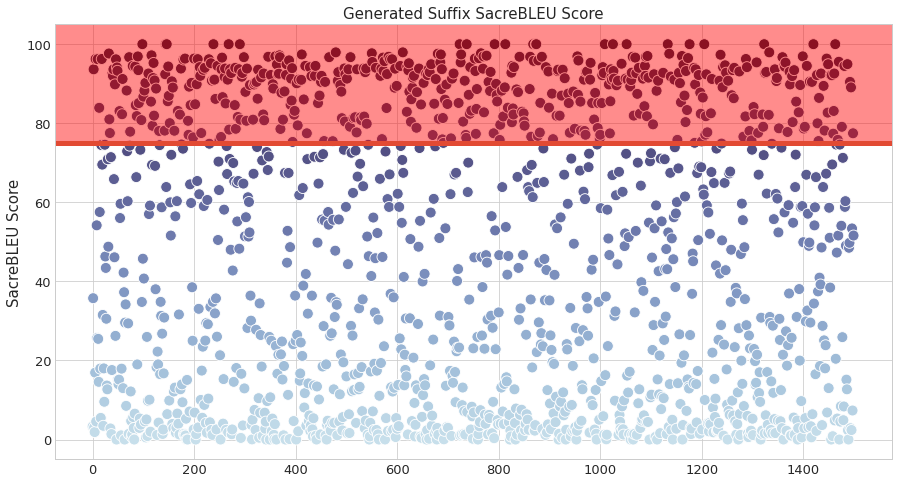}}
    \caption{Generated suffixes Standard Setting} 
    \label{fig7:b} 
    \vspace{4ex}
  \end{subfigure} 
  \begin{subfigure}[b]{0.5\linewidth}
    \centering
    \fbox{\includegraphics[width=0.9\linewidth]{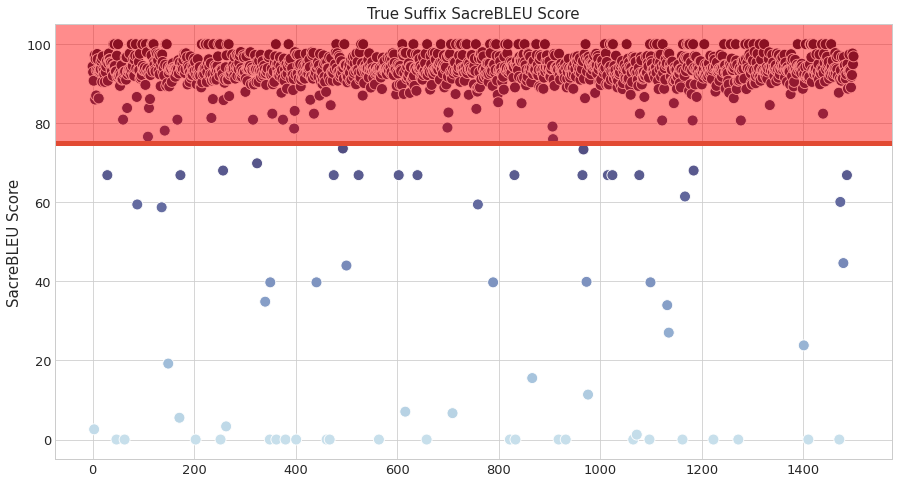}} 
    \caption{True Suffixes Longer Context Setting} 
    \label{fig7:c} 
  \end{subfigure}%%
  \begin{subfigure}[b]{0.5\linewidth}
    \centering
    \fbox{\includegraphics[width=0.9\linewidth]{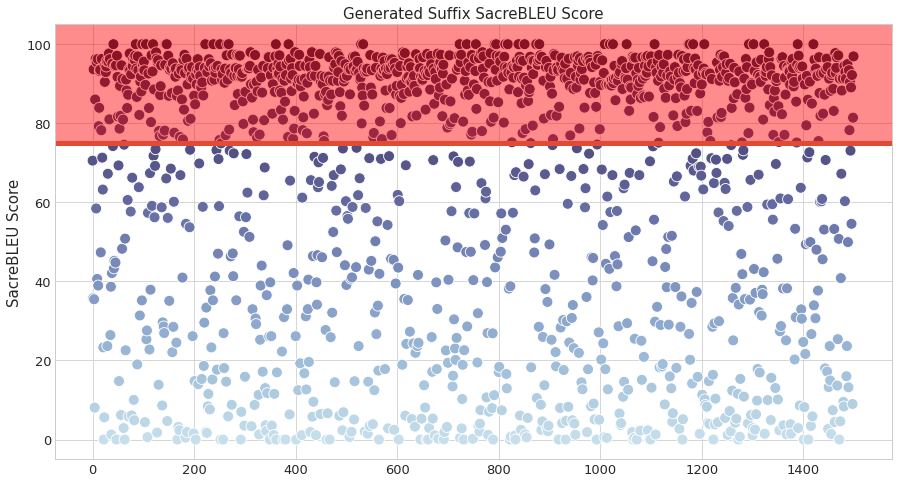}}
    \caption{Generated Suffixes Longer Context Setting} 
    \label{fig:sads} 
  \end{subfigure} 
  
  \caption{Threshold of 75\% Of The True \& Generated Samples SacreBLEU For GPT-Neo 1.3B Standard Setting}
  \label{fig:sp_1.3b}  
\end{figure}
\begin{figure}[ht] 
  \begin{subfigure}[b]{0.5\linewidth}
    \centering
    \fbox{\includegraphics[width=0.9\linewidth]{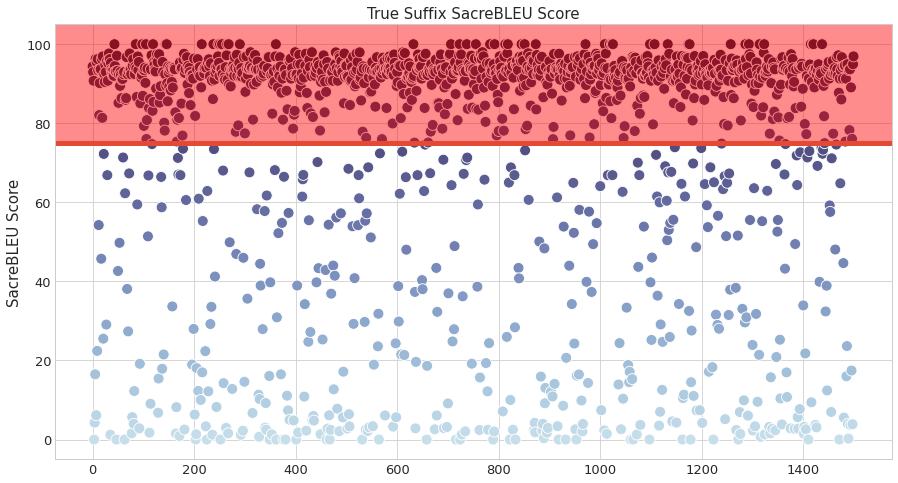}}
    \caption{True Suffixes Standard Setting} 
    \label{fig7:a} 
    \vspace{4ex}
  \end{subfigure}%% 
  \begin{subfigure}[b]{0.5\linewidth}
    \centering
    \fbox{\includegraphics[width=0.9\linewidth]{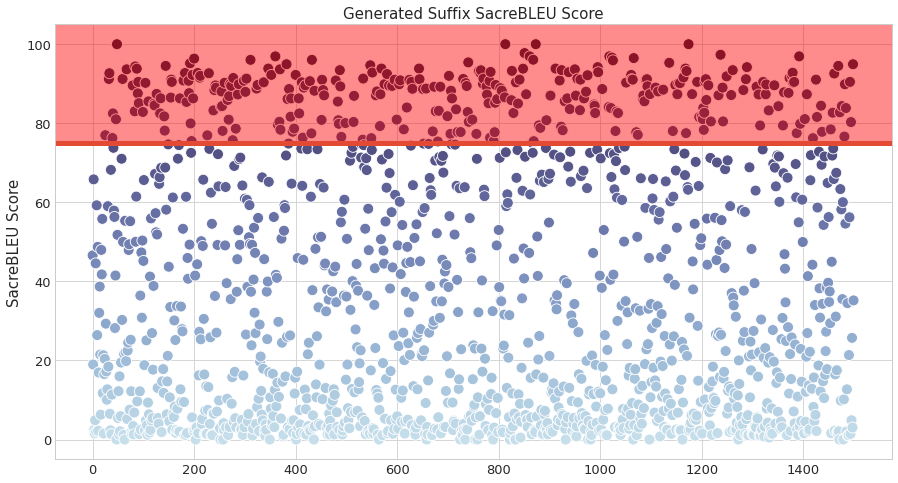}}
    \caption{Generated suffixes Standard Setting} 
    \label{fig7:b} 
    \vspace{4ex}
  \end{subfigure} 
  \begin{subfigure}[b]{0.5\linewidth}
    \centering
    \fbox{\includegraphics[width=0.9\linewidth]{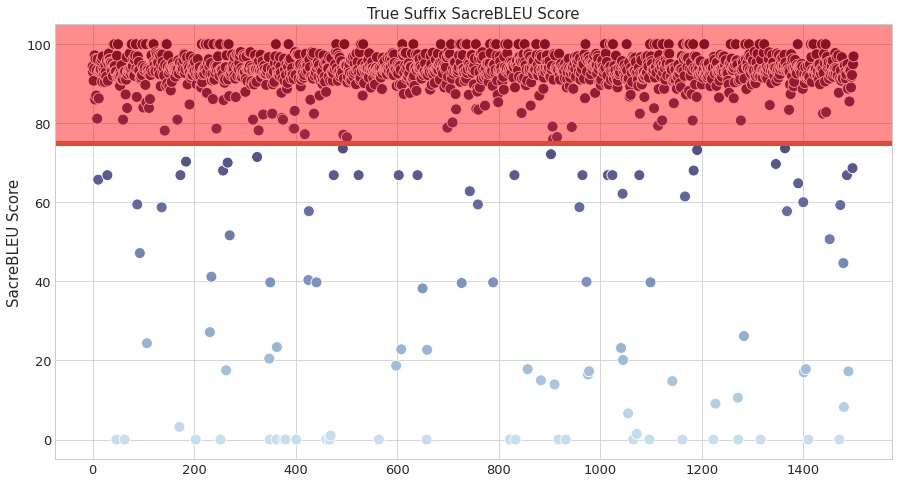}} 
    \caption{True Suffixes Longer Context Setting} 
    \label{fig7:c} 
  \end{subfigure}%%
  \begin{subfigure}[b]{0.5\linewidth}
    \centering
    \fbox{\includegraphics[width=0.9\linewidth]{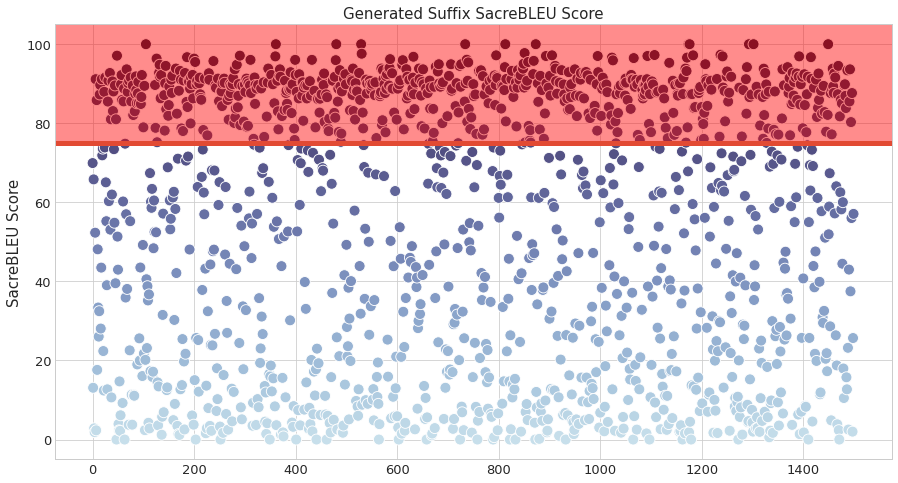}}
    \caption{Generated Suffixes Longer Context Setting} 
    \label{fig:sads} 
  \end{subfigure} 
  \caption{Threshold of 75\% Of The True \& Generated Samples SacreBLEU For GPT-Neo 2.7B Standard Setting}
  \label{fig:sp_2.7b} 
\end{figure}

As shown in \autoref{fig:sp_125m}, the memorization ratio for the GPT-Neo 125M model is relatively low. However, when using standard and longer context settings, there are many instances where the samples are distributed on and beyond the 75\% threshold. Despite this, after implementing the proposed framework, the distribution of samples is more evenly spread across various values rather than being concentrated solely in the region beyond the 75\% threshold.
In contrast to the other variation, GPT-Neo 1.3B \& 2.7B have a large memorization ratio, especially in case of longer context; the framework effect can be seen obviously as many samples exceed the threshold in case of those variations as shown in \autoref{fig:sp_1.3b} and \autoref{fig:sp_2.7b}.

\clearpage
\section{Number Of Epochs \& Reward Functions}
In this section, we will delve deeper into the impact of varying the number of epochs for each reward function across different models. Specifically, we will showcase the results of our experimentation with different settings, such as modifying the reward functions and altering the number of epochs used in each model. This will provide a comprehensive understanding of how the number of training iterations can affect the performance of different models when utilizing different reward functions. As demonstrated in \autoref{table:eps_rf}, there is a strong inverse relationship between the size of the model and the number of epochs required for convergence. As the model size increases, it becomes increasingly efficient at reaching a satisfactory level of performance. This is exemplified by the fact that the smaller variation GPT-Neo 125M requires 16 epochs to converge, while the larger GPT-Neo 2.7B only needs four epochs to achieve an optimal level of performance.

 \begin{table*}[htbp]

\begin{center}
    \centering
    \captionsetup{justification=centering}
\caption{The Results Of Different Reward Functions On The Standard Setting Test Set.}

\label{table:eps_rf}
\small
\resizebox{13cm}{!}{%

\centering
\begin{tabular}{@{}c  |c  |c|  c | c  c @{}}
\toprule

\textbf{Model}& \textbf{Reward Function} & \textbf{Epochs}  & $\bf N-SacreBLEU\uparrow$ & $\bf PPL\downarrow$\\\midrule

\multirow{3}{*}{GPT-Neo 125M} & \multirow{3}{*}{N-SacreBLEU}
 & 8 & 71.77 & 4.43\\
 &   &  12 & 82.87 & 5.25\\
&  & 16 & $87.78 $ & $ 6.55 $
\\ \\ 

\multirow{3}{*}{GPT-Neo 125M} & \multirow{3}{*}{N-SacreBLEU + PPL}
 & 8 & 70.69 & 4.41 \\
 &   &  12 &78.32&4.83\\
&  & 16 & $ 84.35 $ & $ 5.93 $
\\ \\

\multirow{3}{*}{GPT-Neo 125M} & \multirow{3}{*}{BERTScore}
 & 8 & 64.45 & 3.94\\
 &  & 12 & $67.70$ & $4.01$\\  
 &   &  16 & 68.61 & 4.11
\\\midrule
% ---------------------------------------------
\multirow{2}{*}{GPT-Neo 1.3B} & \multirow{2}{*}{N-SacreBLEU}
 & 8 & 72.10 & 3.65\\
 &   &  12 & 85.53 & 6.43
\\ \\ 

\multirow{2}{*}{GPT-Neo 1.3B} & \multirow{2}{*}{N-SacreBLEU + PPL}
 & 8 & 52.02 & 2.85\\
 &   &  12 &52.05&2.86\\
\\ 

\multirow{3}{*}{GPT-Neo 1.3B} & \multirow{3}{*}{BERTScore}
 & 8 & 53.22 & 2.64\\
 &   &  12 & 57.01 & 2.77\\
&  & 16 & $68.47$ & $2.95$
\\\midrule

% ---------------------------------------------
\multirow{2}{*}{GPT-Neo 2.7B} & \multirow{2}{*}{N-SacreBLEU}
 & 4 & 71.47 & 6.10\\
&  & 8 & 83.09 & 8.423\\
\\ 

\multirow{2}{*}{GPT-Neo 2.7B} & \multirow{2}{*}{N-SacreBLEU + PPL}
 &4 & 80.84 & 8.60\\
 &  & 8 & 88.67 & 12.53\\
\\

\multirow{2}{*}{GPT-Neo 2.7B} & \multirow{2}{*}{BERTScore}
 &4 & 61.79 & 4.46\\
 &  & 8 & 75.09 & 6.76
\\\midrule

\end{tabular}%
}
\normalsize
\end{center}
\end{table*}

\clearpage

\section{Qualitative Results}
\begin{figure}[htbp]
    \centering
    \includegraphics[width=1\linewidth]{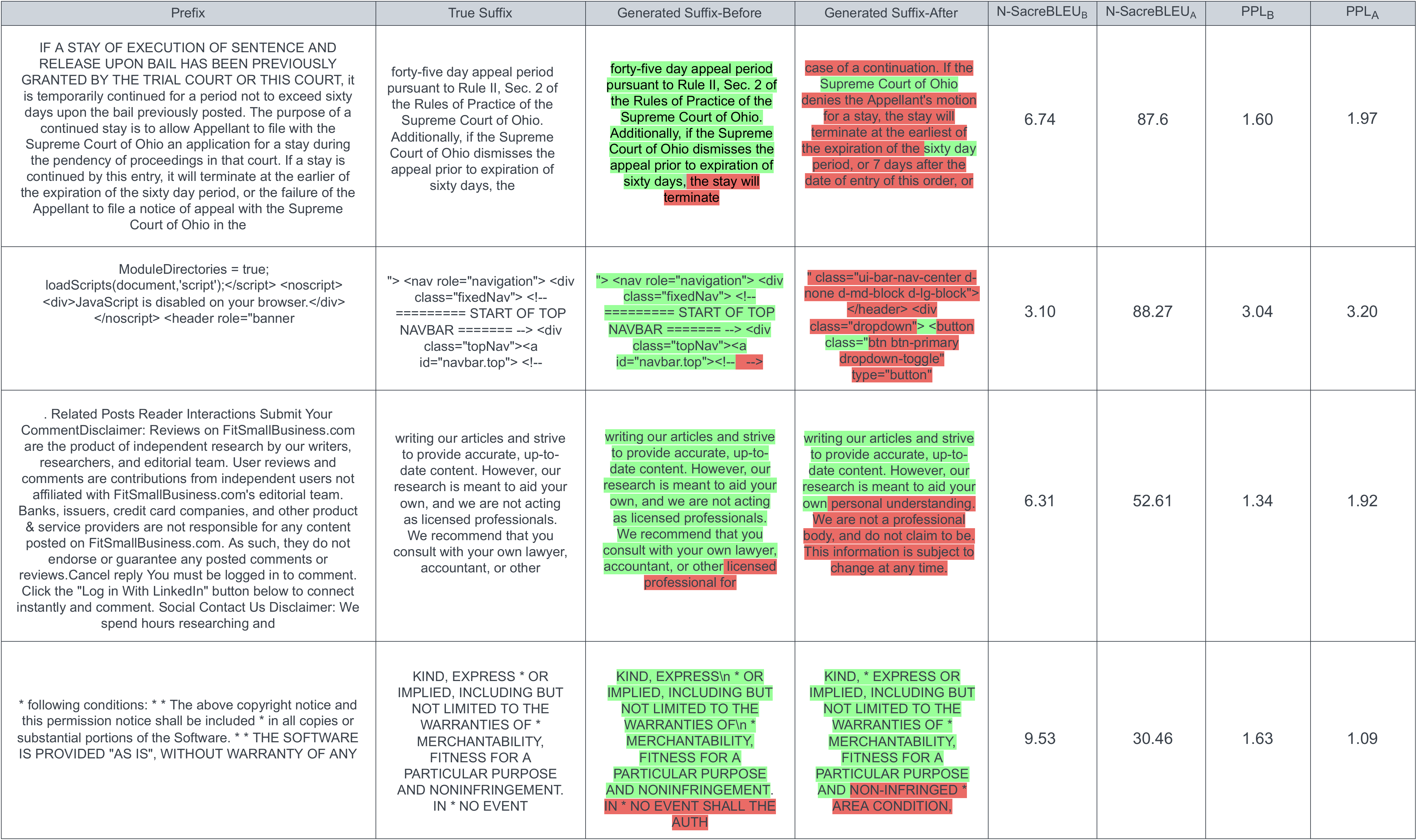}
    \caption{Suffixes that are memorized by the employed language models and the generated suffixes given the same prefix. Green indicates that this part is memorized according to the true suffix, while red indicates that it’s dissimilar.}
    \label{fig:mem_results}
\end{figure}

\begin{figure}[htbp]
    \centering
    \includegraphics[width=1\linewidth]{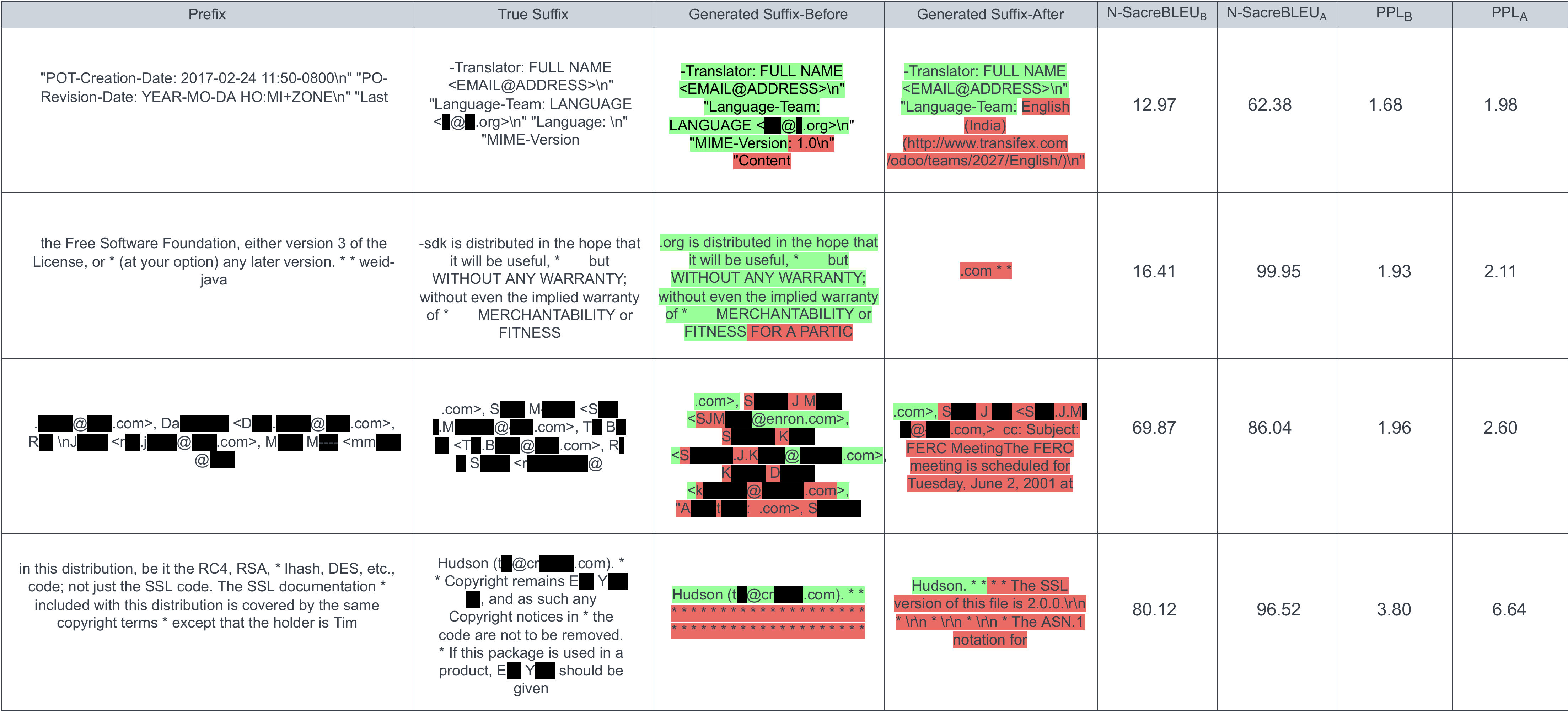}
    \caption{Suffixes that are memorized by the employed language models and the generated suffixes given the same prefix. Green indicates that this part is memorized according to the true suffix, while red indicates that it’s dissimilar.}
    \label{fig:mem_results_two}
\end{figure}

In this section, we demonstrate the effectiveness of our proposed framework by presenting a thorough analysis of samples generated before and after its application. To provide a comprehensive evaluation, we have chosen samples from various model sizes, including 125M, 1.3B, and 2.7B, and included examples from both standard and longer contexts. Additionally, we present samples from different training phases to showcase the learned policy's evolution over time. As previously mentioned, the policy initially focuses on replacing individual words or numbers to decrease the similarity between samples. As the training process progresses, the policy becomes more aggressive and replaces entire phrases, as shown in \autoref{fig:mem_results}.
\autoref{fig:mem_results_two} demonstrates that the framework is capable of learning a policy that reduces or eliminates the amount of memorized personal data, such as email addresses. However, it should be noted that in certain instances, this can increase perplexity.

\clearpage

\section{Median Comparison}

\begin{figure}[htb]
    \centering % <-- added
\begin{subfigure}{0.31\textwidth}
  \fbox{\includegraphics[width=1\linewidth]{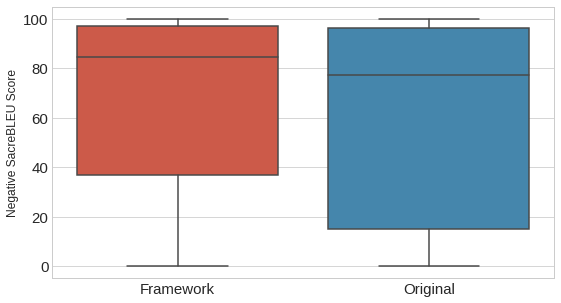}}
  \caption{GPT-Neo 125M On Standard Setting}
  \label{fig:1}
\end{subfigure}\hfil % <-- added
\begin{subfigure}{0.31\textwidth}
  \fbox{\includegraphics[width=\linewidth]{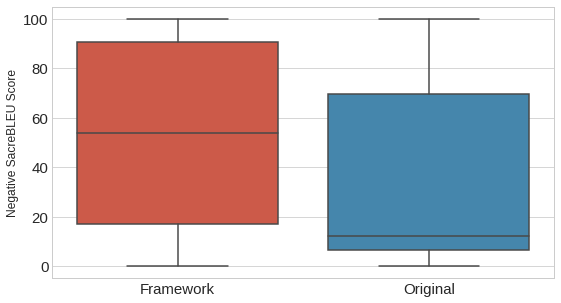}}
  \caption{GPT-Neo 1.3B On Standard Setting}
  \label{fig:2}
\end{subfigure}\hfil % <-- added
\begin{subfigure}{0.31\textwidth}
  \fbox{\includegraphics[width=\linewidth]{Figures/bp2.7Stand.png}}
  \caption{GPT-Neo 2.7B On Standard Setting}
  \label{fig:3}
\end{subfigure}

\medskip
\begin{subfigure}{0.31\textwidth}
  \fbox{\includegraphics[width=\linewidth]{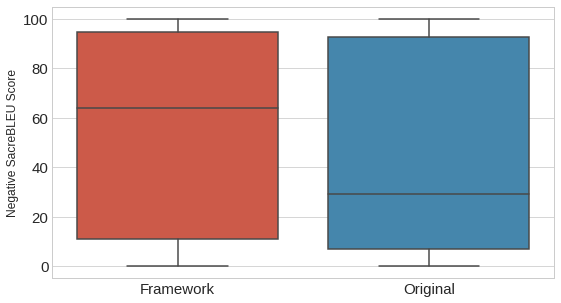}}
  \caption{GPT-Neo 125M On Long Setting}
  \label{fig:4}
\end{subfigure}\hfil % <-- added
\begin{subfigure}{0.31\textwidth}
  \fbox{\includegraphics[width=\linewidth]{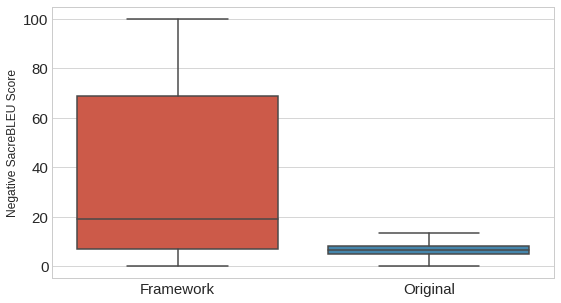}}
  \caption{GPT-Neo 1.3B On Long Setting}
  \label{fig:5}
\end{subfigure}\hfil % <-- added
\begin{subfigure}{0.31\textwidth}
  \fbox{\includegraphics[width=\linewidth]{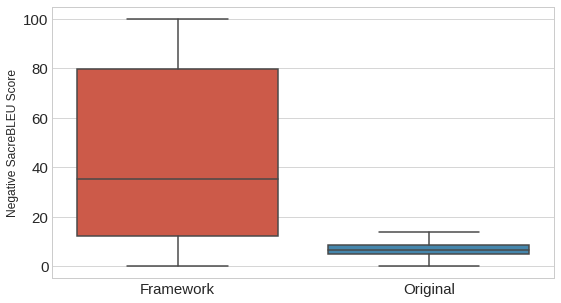}}
  \caption{GPT-Neo 2.7B On Long Setting}
  \label{fig:6}
\end{subfigure}
\caption{Displaying The Negative SacreBLEU Distribution of The Models On Standard \& Long Settings Before \textcolor{blue}{(blue)} \& After \textcolor{orange}{(orange)} Applying The Framework}
\label{fig:images}
\end{figure}

\section{Computational Resources}
In order to fine-tune GPT-Neo models of sizes 125M and 1.3B, we utilized a cluster of two V100 GPUs, each equipped with 32GB of VRAM. The 125M model required approximately 0.38 minutes per PPO epoch, resulting in a total computation time of 3.04 minutes for six epochs. The 1.3B model required a slightly longer computation time of 1.68 minutes per PPO epoch, for a total of 13.44 minutes over eight epochs. For the largest variant, GPT-Neo 2.7B, we utilized a cluster of four V100 GPUs, each with 32GB of VRAM, and employed a sharding strategy with zero 3 \cite{rasley2020deepspeed}. Each PPO epoch for this model required 5.125 minutes, resulting in a total computation time of approximately 20 minutes over four epochs. For finetuning those models, we employed the HuggingFace library \cite{wolf-etal-2020-transformers} for training and Pytorch \cite{paszke2017automatic} for parallelizing the model.

\end{document}